\pdfoutput=1

\documentclass[11pt]{article}

\usepackage{acl}

\usepackage{times}
\usepackage{latexsym}

\usepackage[T1]{fontenc}

\usepackage[utf8]{inputenc}

\usepackage{microtype}

\usepackage{inconsolata}

\usepackage{amsmath}
\usepackage{amsfonts}
\usepackage{amssymb}
\usepackage{bbm}
\usepackage{enumitem}
\usepackage{multirow}
\usepackage{booktabs}
\usepackage{graphicx}
\usepackage{float}
\usepackage{subfigure}
\usepackage{booktabs}
\usepackage{subcaption}
\usepackage[normalem]{ulem}
\useunder{\uline}{\ul}{}

\title{F-MALLOC: Feed-forward Memory Allocation for Continual Learning in Neural Machine Translation}

\author{Junhong Wu, Yuchen Liu, Chengqing Zong \\
        \footnotesize School of Artificial Intelligence, University of
        Chinese Academy of Sciences, Beijing, China \\ 
        \footnotesize State Key Laboratory of Multimodal Artificial Intelligence Systems, Institute of Automation, CAS, Beijing, China \\
        \footnotesize{wujunhong2021@ia.ac.cn;\{yuchen.liu,cqzong\}@nlpr.ia.ac.cn
        }
}

\begin{document}
\maketitle
\begin{abstract}
In the evolving landscape of Neural Machine Translation (NMT), the pretrain-then-finetune paradigm has yielded impressive results. However, the persistent challenge of Catastrophic Forgetting (CF) remains a hurdle. While previous work has introduced Continual Learning (CL) methods to address CF, these approaches grapple with the delicate balance between avoiding forgetting and maintaining system extensibility. To address this, we propose a CL method, named \textbf{F-MALLOC} (\textbf{F}eed-forward \textbf{M}emory \textbf{ALLOC}ation). F-MALLOC is inspired by recent insights highlighting that feed-forward layers emulate neural memories and encapsulate crucial translation knowledge. It decomposes feed-forward layers into discrete memory cells and allocates these memories to different tasks. By learning to allocate and safeguard these memories, our method effectively alleviates CF while ensuring robust extendability. Besides, we propose a comprehensive assessment protocol for multi-stage CL of NMT systems. Experiments conducted following this new protocol showcase the superior performance of F-MALLOC, evidenced by higher BLEU scores and almost zero forgetting.\footnote{The code and data for this work are available at \url{https://github.com/WJMacro/ContinualMT}} 
\end{abstract}

\section{Introduction}

In the pursuit of achieving state-of-the-art results in Neural Machine Translation (NMT), the reliance on large-scale parallel corpora has been pivotal \citep{DBLP:journals/corr/BahdanauCB14, DBLP:conf/nips/VaswaniSPUJGKP17}. However, practical application scenarios often present challenges, especially when translation is necessitated for specific domains with limited data resources \citep{DBLP:journals/corr/abs-1806-00258, DBLP:journals/jair/Saunders22}. Typically, the prevalent paradigm involves the initial pretraining of models on expansive general domain corpus, followed by finetuning for the target domain \citep{DBLP:journals/corr/FreitagA16, DBLP:journals/corr/abs-1906-07978}.

Despite the efficacy of this pretrain-then-finetune paradigm, it has been demonstrated that fine-tuning on the target domain can result in significant performance degradation in the general domain, a phenomenon known as Catastrophic Forgetting (CF) \citep{DBLP:conf/nips/French93}. In response to this challenge, various Continual Learning (CL) approaches have emerged to address CF in NMT systems. Existing efforts primarily rely on regularization-based techniques to constrain the divergence of model parameters from their previous values \citep{DBLP:conf/aclnmt/KhayrallahTDK18, DBLP:conf/acl/SaundersSGB19,DBLP:conf/naacl/CaoWCW21}. While these methods are mathematically elegant, they still face challenges related to forgetting.  Alternatively, some approaches take an architecture-based framework, isolating parameters specific to different tasks to prevent forgetting \citep{DBLP:conf/naacl/GuFX21, DBLP:conf/aaai/LiangZWQ021, DBLP:conf/acl/HuangLMYL23}. 
However, they require prior information on task numbers to allocate parameters and rely on external storage of model or mask matrices, limiting its extendibility and applicability.



In summary, the demand for a CL method for NMT systems that is both extendable and effective in preventing forgetting is pressing. To this end, we introduce a new CL method termed \textbf{F-MALLOC} (\textbf{F}eed-forward \textbf{M}emory \textbf{ALLOC}ation),
which is inspired by recent insights that feed-forward layers emulate neural memories and 
encapsulate crucial translation knowledge 
\citep{DBLP:conf/emnlp/GevaSBL21,DBLP:conf/acl/HuangLMYL23}. Therefore, we facilitate new knowledge learning and mitigate CF by allocating and protecting these memories. F-MALLOC first leverages a structural pruning method to trim the feed-forward layers of a pretrained NMT model, preserving memories that encapsulate crucial general domain knowledge. Subsequently, F-MALLOC proceeds to learn a set of non-exclusive task masks \citep{DBLP:conf/icml/SerraSMK18}, automatically allocating the `writable' memory capacity to upcoming tasks. The memories allocated in this manner are then designated as `read-only'. F-MALLOC strategically blocks gradient flows through these `read-only' memories, effectively mitigating the risk of forgetting. 

Meanwhile, conventional CL evaluation protocols in the NMT area typically focus on a single stage of training, lacking a holistic perspective over multiple stages. 
Therefore, we introduce a comprehensive evaluation protocol for multi-stage CL in the NMT scenario. 
Our protocol incorporates metrics assessing forgetting mitigation and adaptation to novel tasks. To enhance robustness, we conduct tests with random task sequences, reducing biases from specific orders. This protocol provides a nuanced understanding of F-MALLOC and competing methods' performance over time in NMT.

Experiments conducted following the proposed protocol highlight the superior performance of F-MALLOC with high robustness. Additional analysis of F-MALLOC's memory allocation strategy reveals its effective utilization of task information, such as inherent difficulty or inter-task similarities, resulting in enhanced performance.

In summary, the contributions of this paper are as follows:
\begin{itemize}
    \item We propose F-MALLOC, a multi-stage CL method that prevents forgetting and promotes new knowledge acquisition through feed-forward memory allocation. It requires no prior task information and minimal storage overhead.
    \item Through a tailored evaluation protocol for multi-stage CL in NMT systems, we enhance the understanding of system performance on both stability and plasticity.
    \item Further analysis of F-MALLOC's adaptive memory allocation strategy demonstrates its effectiveness in leveraging task difficulty and inter-task similarities to optimize capacity usage and encourage knowledge transfer.
\end{itemize}
\section{Background}

\begin{figure*}[ht]
\centering
\includegraphics[width=0.95\textwidth]{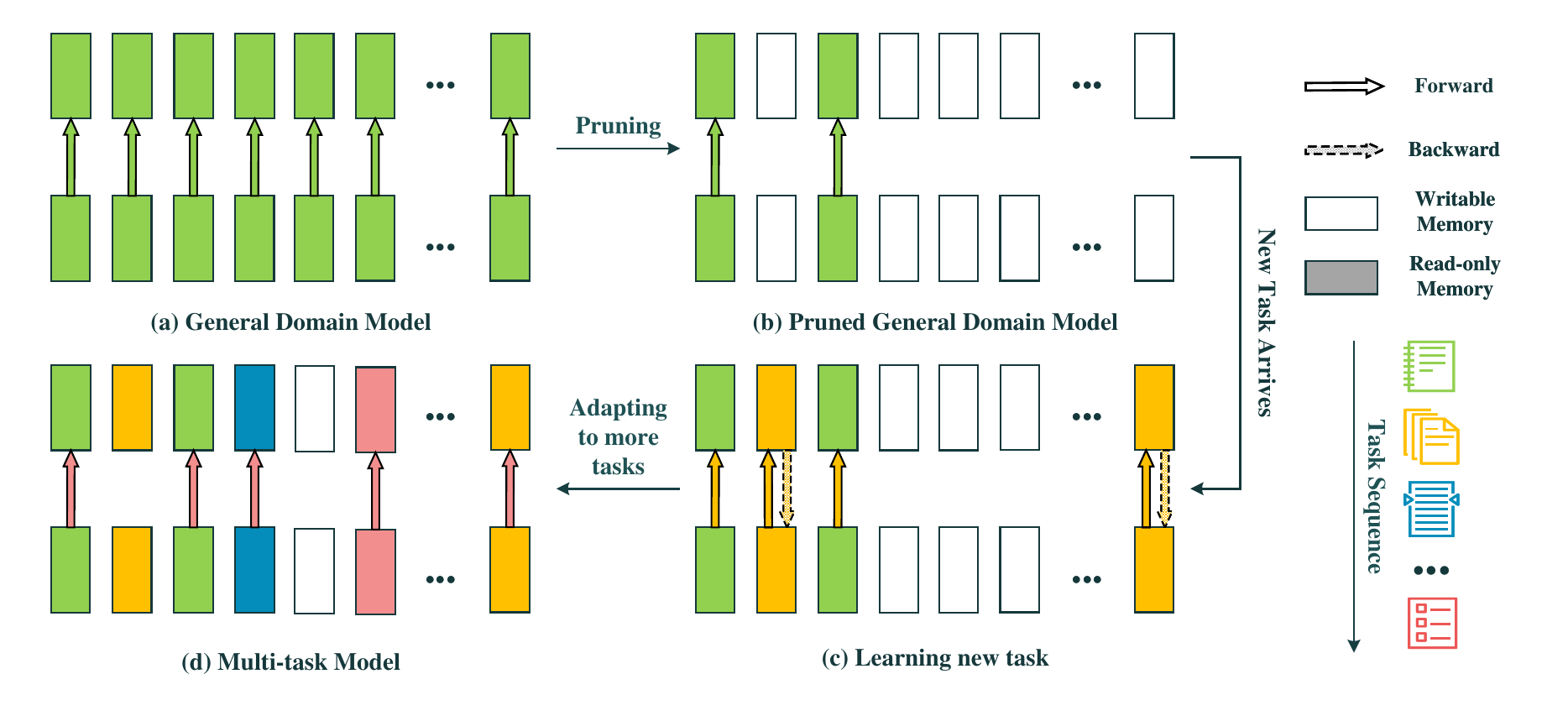}
\caption{Illustration of F-MALLOC. For simplification, we depict a decomposed feed-forward layer. (a) \textbf{The Original General Domain Model:} Highlighting the general domain task in green. (b) \textbf{Pruned General Domain Model:} Post-pruning, pruned memories are `writable' (depicted in white), while others are designated as `read-only.' (c) \textbf{Learning a New Task:} The model learns to allocate some memories to the new task and mark them `read-only' (depicted in yellow). `read-only' memories remain available for future tasks' forward propagation. However, backward propagation through them is prohibited. (d) \textbf{Multi-task Model:} After learning all tasks, each task occupies a share of memory capacity. The forward pass of the last task is shown.}
\label{Fig:overview}
\end{figure*}

\subsection{Feed-forward Layers Emulate Memory Networks}

\paragraph{Feed-forward Layer.} The prevalent architecture in NMT is the encoder-decoder Transformers\citep{DBLP:conf/nips/VaswaniSPUJGKP17}, which is made of intertwined multi-head attention (MHA) and point-wise feed-forward layers. Our specific focus lies in the feed-forward layer, formally defined as:
\begin{equation}\small
    \operatorname{FF}(x) = W^{(2)} \cdot \sigma(W^{(1)} \cdot x)\
    \label{Eq.1:feed-forward-memory}
\end{equation}
where $W^{(1)}, W^{(2)}$ represent learnable parameters (bias term omitted for simplification), and $\sigma$ typically denotes the activation function, commonly ReLU.


\paragraph{Feed-forward layer as neural memory of knowledge.}
Recent research has explored the interpretability of feed-forward structures, noting a significant resemblance between the feed-forward layer and neural memory \citep{DBLP:conf/nips/SukhbaatarSWF15}. Treating parameter matrices $W^{(1)}$ and $W^{(2)}$ as keys and values respectively, the feed-forward layer can be seen as an unnormalized key-value memory \citep{DBLP:journals/corr/abs-1907-01470}. Studies have delved into this similarity, with \citet{DBLP:conf/emnlp/GevaSBL21} revealing that in feed-forward layers, each key correlates with textual patterns in training examples, while each value induces a distribution over the output vocabulary. In the context of Neural Machine Translation, \citet{DBLP:conf/acl/HuangLMYL23} demonstrates that feed-forward layers encapsulate crucial translation knowledge and can facilitate knowledge transfer between models.

\section{Methods}
\subsection{Overview}
Building upon prior research that characterizes feed-forward layers as neural memory repositories of knowledge, we posit a hypothesis that effective allocation and protection of these memories within feed-forward layers can facilitate both the acquisition of new knowledge and the prevention of forgetting. Our proposed method, F-MALLOC, is devised on the premise of this hypothesis.


F-MALLOC is specifically tailored to the feed-forward structure, with all other parameters held constant throughout the process. To preserve critical general domain knowledge while allowing flexibility for future task learning, we initiate the process with a structured pruning method (\ref{Pruning}). This method aids in eliminating unimportant memories, making them `writable’. Subsequently, we introduce learnable task masks to manage these free memories (\ref{ContinualLearning}). These task masks, acquired through learning, play a vital role in memory allocation for new tasks, designating them as `read-only' to prevent alterations. For an overview of our method, please refer to Fig.\ref{Fig:overview}.

\subsection{Preserving General Domain Knowledge}
\label{Pruning}
Pruning has demonstrated effectiveness in retaining essential parameters while eliminating unnecessary ones in neural networks. In this context, we adopt a structured pruning method, which is designed to preserve general domain knowledge. The pruning process calculates an importance score for each memory cell in the feed-forward layer, retaining only the most crucial ones. 


\paragraph{Importance-based memory pruning.} The pruning problem can be seen as finding an optimal mask under a sparsity constraint. To formalize this, we decompose the feed-forward layer into $N$ key-value pairs, which we call a memory cell\footnote{We use memory and memory cell interchangeably.}. Subsequently, a mask is introduced to control the activation of them \citep{DBLP:conf/acl/XiaZC22,DBLP:conf/nips/KwonKMHKG22}:
\begin{equation}\small
    \operatorname{FF}(x,m) = \sum_{i=1}^{N}m_i \odot W_{:,i}^{(2)} \cdot \sigma(W_{i,:}^{(1)} \cdot x)
    \label{Eq.3:feed-forward-mask}
\end{equation}
Here, $N$ denotes the hidden dimension of the feed-forward layer, $m \in \{0,1\}^{N}$ represents the mask vector and $\odot$ denotes the Hadamard product.

A common approach to select unnecessary memory is to estimate the importance of different memories with gradient \citep{DBLP:conf/nips/MichelLN19} or Fisher information \citep{DBLP:conf/icml/LiuZKZXWCYLZ21}. The precise calculation of the importance score typically demands the use of the same data and loss functions employed during model training, which is often impractical in CL scenarios where obtaining the training data of a pretrained model may be unfeasible. 

\begin{figure}[ht]
\centering
\includegraphics[width=0.50\textwidth]{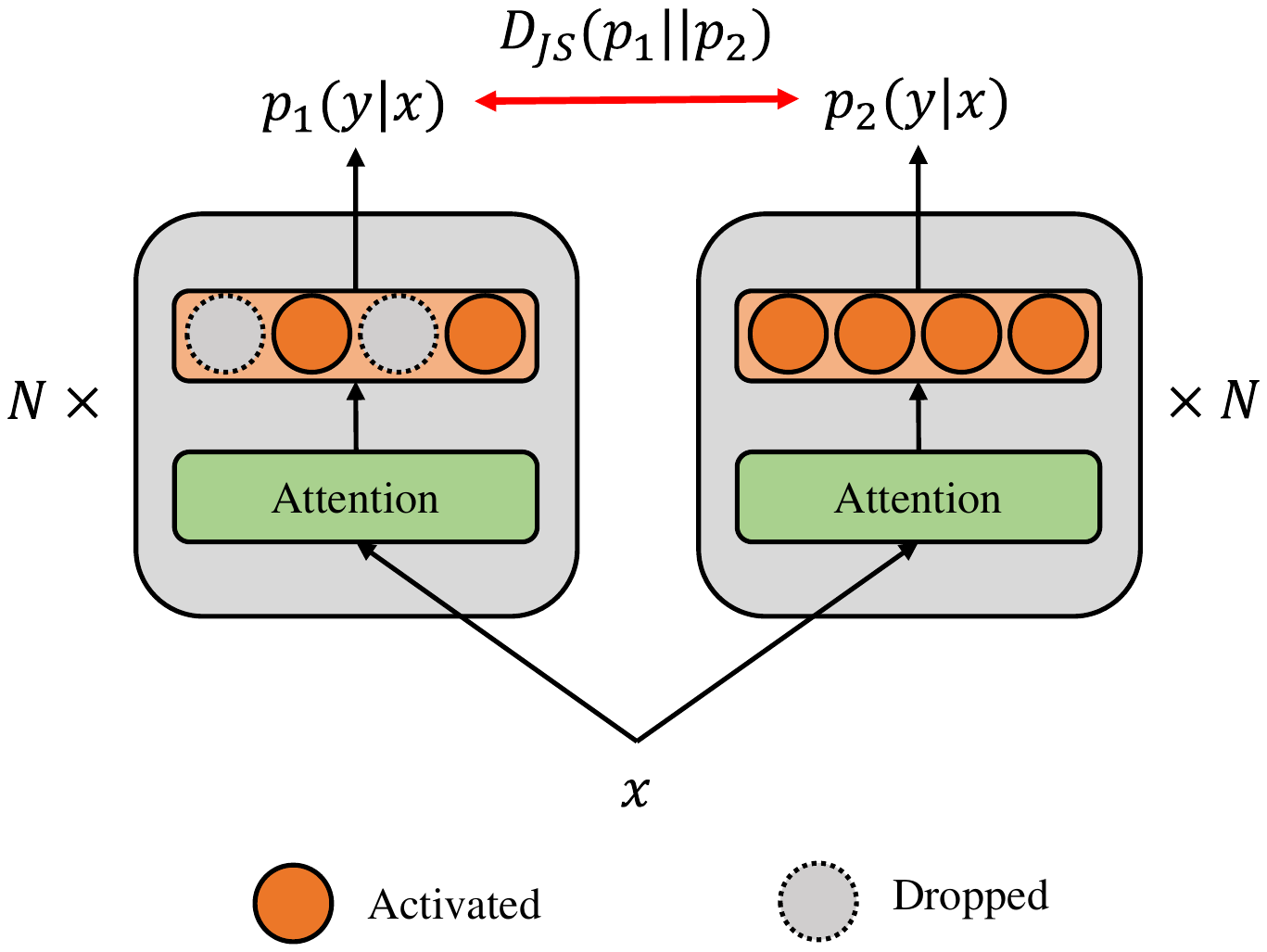}
\caption{Illustration of estimating feed-forward memory importance via JS divergence.}
\label{Fig:js_loss}
\end{figure}

To address this challenge, we propose an alternative approach employing Jensen–Shannon (JS) divergence. The method draws inspiration from the stochastic dropout mechanism \citep{DBLP:journals/corr/abs-1207-0580}, which introduces randomness by eliminating a portion of units in each layer during training, mitigating co-adaptation and overfitting. In our approach, dropout is applied to the feed-forward layer, generating unique memory activations and distinct outputs during each forward pass. By comparing these outputs and computing the gradient of the divergence, we derive a novel importance score for memories.

As shown in Fig.\ref{Fig:js_loss}, we perform two forward passes, with or without dropout, of the input data $x$ through the network, generating two distributions of model predictions, denoted as $\mathcal{P}_1(y|x)$ and $\mathcal{P}_2(y|x)$. We then calculate the JS divergence between these predictions:
\begin{equation}\small
\begin{aligned}
\mathcal{L}_{\operatorname{JS}}(x) = \frac{1}{2}( & \operatorname{KL}(\mathcal{P}_1(y|x),\mathcal{P}_2(y|x)) \\
& + \operatorname{KL}(\mathcal{P}_2(y|x),\mathcal{P}_1(y|x)))
\end{aligned}
\label{Eq.5:JS}
\end{equation}
where $\operatorname{KL}(\cdot,\cdot)$ denotes the Kullback–Leibler (KL) divergence. 
In practice, we adopt an external dataset $\mathcal{D}$ and estimate the average gradient of JS divergence, serving as an empirical importance score:
\begin{equation}\small
    I_k = \mathbb{E}_{x \in \mathcal{D}} \left|\frac{\partial \mathcal{L}_{\operatorname{JS}}(x)}{\partial m_k}\right|
    \label{Eq.4:Importance}
\end{equation}

Following the derivation of the importance score, a binary mask is generated through a binarization function utilizing the $s$ quantile of the importance score, denoted as $q_s(I)$, as the threshold:
\begin{equation}\small
m_k^G= \begin{cases}1, & \text { if } I_k \geq q_s(I) \\ 0, & \text { if } I_k<q_s(I)\end{cases}
\label{Eq.7:gen-mask}
\end{equation}
where $s$ is the desired sparsity. Substituting this mask $m^G$ into Formula \ref{Eq.3:feed-forward-mask} accomplishes the pruning.

\subsection{Learning New Domain Continually}
\label{ContinualLearning}
\begin{figure}[ht]
\centering
\includegraphics[width=0.50\textwidth]{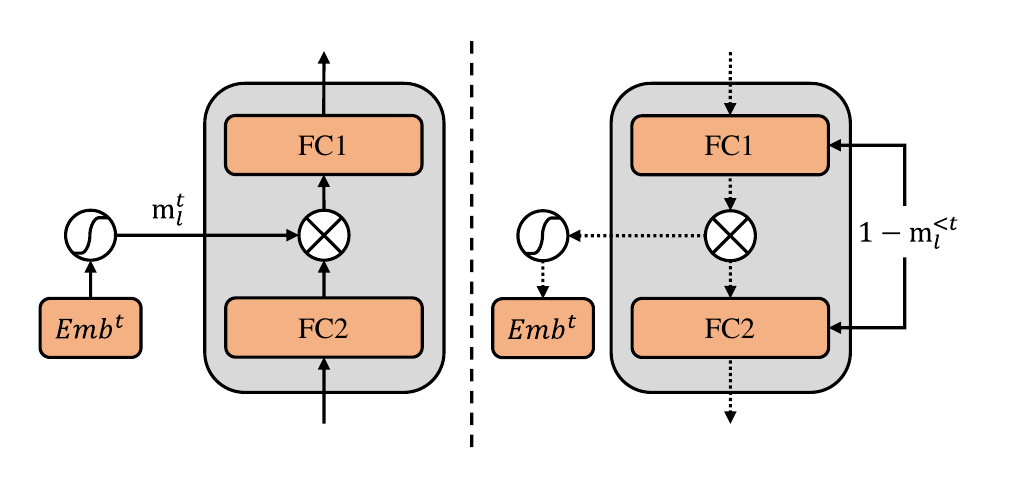}
\caption{Illustration of new domain learning: forward (the left) and backward (the right) propagation. Here, we show the inner structure of the feed-forward layer.}
\label{Fig:new_domain}
\end{figure}

After the structure pruning stage, wherein specific feed-forward memories are pruned and marked `writable' for future learning, we introduce a task mask mechanism to manage memory. Throughout the forward pass, these task masks govern the activation of feed-forward memory, conditioning the model for specific tasks. In the backward pass, the task masks are employed to suppress gradient updates to the `read-only' memories, effectively preventing CF. Fig.\ref{Fig:new_domain} provides an overview of this procedure.

\paragraph{Learning task mask to allocate `writable' memory.}
To adapt to a new task $t$, a task mask $m_l^t$ is learned. This task mask serves to conditionally activate the memories in the $l$-th feed-forward layer. We adopt the task-based hard attention mechanism proposed by \citet{DBLP:conf/icml/SerraSMK18} to train the mask. For each task $t$, a learnable task embedding $e_l^t$ is introduced for each layer. The task mask $m_l^t$ is defined as a gated version of the embedding vector $e_l^t$:
\begin{equation}\small
    m_l^t = \sigma(\frac{e_l^t}{\tau})
\label{Eq.11:TaskMask}
\end{equation}
where $\sigma$ represents a gate function, and $\tau$ is a temperature variable. We wish to learn a binary task mask that could be employed to allocate feed-forward memory in the same format as described in Eq.\ref{Eq.3:feed-forward-mask}. 

To facilitate the efficient learning of task masks, we employ a sigmoid function with a temperature scalar to create a differentiable pseudo-gate function. The temperature scalar regulates the polarization or `hardness' of the pseudo-step function. As $\tau \to 0$, the values of $m_{l, i}^t$ tend towards either $0$ or $1$, compelling the model to exploit allocated memories. Conversely, as $\tau \to \infty$, the values of $m_{l, i}^t$ approach $0.5$, allowing the model to freely explore memories. Throughout the training process, we implement temperature annealing, transitioning from $\frac{1}{\tau_{max}}$ to $\tau_{max}$. This dynamic adjustment aids the model in cyclically exploring memories while simultaneously exploiting activated memories. During the training process, the mask undergoes a gradual polarization, resulting in the occupation of useful memories. 

The embedding is initialized with $\alpha m^G-|\mathcal{N}(0,1)|$, where $\alpha$ is a constant. This initialization enables the new domain to harness all the insights from the general domain's memories. Simultaneously, it initializes the excess capacity usage to zero at the outset of training, fostering minimal capacity utilization. Upon model convergence, we archive the acquired mask for future utilization.

\paragraph{Applying task mask to safeguard `read-only' memory.} To tackle the challenge of CF, our approach involves leveraging task masks acquired from previous tasks to influence the gradient. Before learning a new task, denoted as $t$, we aggregate all task masks from preceding tasks using an element-wise max (EMAX) operation and subsequently binarize the result with a threshold $\lambda$, as expressed by the following equation:
\begin{equation}\small
    m_l^{<t} = \operatorname{I}_\lambda(\operatorname{EMAX}_{j<t} \{m_l^j\})
\label{Eq.12:PreviousMask}
\end{equation}
Here, the subscript $l$ denotes the layer index. In this specific context, task 0 corresponds to the general domain translation task, and the associated mask derived from structural pruning for the general domain is denoted as $m^0$. The aggregated binary mask $m_l^{<t}$ encapsulates critical memories designated as `read-only' by previous tasks. The primary objective is to safeguard the parameters in these memories, preserving their functionality for previous tasks. To achieve this, we utilize the mask to adjust the gradient during the training of task $t$, as articulated in the following equation:
\begin{equation}\small\small
    g_l^{'t}= g_l^t \odot (1 - m_l^{<t})
\label{Eq.13:GradientMask}
\end{equation}
where $\odot$ denotes the Hadamard product. This modification guarantees that memories crucial for previous tasks (entries with a value of 1 in $m_l^{<t}$) will have near zero gradients, thereby ensuring their preservation during the training of subsequent tasks.


\section{Experiments}

In our multi-stage CL experiments for NMT systems, we finetune a pretrained general domain model on $T$ new domains successively\footnote{In our experiments, a task is a domain. Hence, we use task and domain interchangeably.}. The pretrained model is based on the WMT’19 German-English news translation task winner \citep{DBLP:conf/wmt/NgYBOAE19}. To neutralize the impact of task order, we randomly generate five task order sequences and report the average result.

\subsection{Data Preparation}
In the context of structure pruning, we employ the WMT14 de-en translation data\footnote{https://www.statmt.org/wmt14/translation-task.html} as the external dataset. Additionally, we combine the WMT newstest datasets from 2019 to 2021\footnote{https://www.statmt.org/} to form a comprehensive general domain test set.
For the continual domain adaptation experiments, we utilize the OPUS multi-domains dataset \citep{DBLP:conf/aclnmt/KoehnK17}, which has been re-split by \citet{DBLP:conf/acl/AharoniG20}. It includes German-English parallel data in five domains: Medical, Law, IT, Koran and Subtitles.

The details of all datasets mentioned above are shown in Appendix \ref{Appendix:dataset}.

\subsection{Baseline and Implementation Details}
\paragraph{Baseline systems.} We incorporate eight competitive methods for comparison in our experiments, which can be categorized into two groups: Non-Continual Learning (Non-CL) methods and CL methods. In the Non-CL category, \textbf{(1) Single-domain} and \textbf{(2) Mixed-domain} directly finetune the pretrained model on single or mixed domain data, achieving the \textbf{upper bound} performance. \textbf{(3) Adapter \citep{DBLP:conf/emnlp/BapnaF19}} inserts Adapters on each transformer block of the general domain model. In the CL category, we have \textbf{(4) Sequential Fine-tuning} continual finetunes the pretrained model sequentially; \textbf{(5) EWC \citep{DBLP:conf/naacl/ThompsonGKDK19,DBLP:conf/acl/SaundersSGB19}} adds elastic weight consolidation term to regularize loss; \textbf{(6) KD\citep{DBLP:conf/aclnmt/KhayrallahTDK18,DBLP:conf/mtsummit/DakwaleM17}} use knowledge distillation to transfer knowledge; \textbf{(7) Dynamic-KD\citep{DBLP:conf/naacl/CaoWCW21}} involves dynamic adjustments to the weight of KD loss. \textbf{(8) PTE\citep{DBLP:conf/naacl/GuFX21}} prune the general domain model and learn target domain with free parameters. We have extended the baseline method designed for a single stage to multiple stages. Further details on these methods can be found in Appendix \ref{Appendix:baseline}.

\paragraph{Implementation detail.} In our proposed method, we exclusively finetune the Feed-forward layers in Transformers, keeping all other modules frozen throughout the procedure. During the structure pruning stage, we set the pruning sparsity to 0.2 for subsequent CL experiments (the same pruning sparsity is also used in PTE for fair comparison). When computing the proposed JS loss, we only activate the default dropout module for feed-forward activation. In the CL stage, the temperature hyperparameter $\tau_{max}$ is set to 400, following previous work \citep{DBLP:conf/icml/SerraSMK18}. We use $\alpha=5.0$ in the embedding initialization. The binarize threshold $\lambda$ in Eq.\ref{Eq.12:PreviousMask} is set to 0.5. For more details please refer to Appendix \ref{Appendix:implementation_details}.

\subsection{Metrics}
We adopt the BLEU score to evaluate the translation performance. Recognizing that post-training BLEU may not sufficiently capture the nuances in multi-stage CL, we introduce two additional metrics: \textbf{Forgetting Ratio (FR)} and \textbf{Saturation Ratio (SR)}.
\begin{itemize}
    \item Inspired by \citep{DBLP:conf/cvpr/LiuSLSS20}, FR is defined as:
    \begin{equation}\small
    \operatorname{FR}^t = \frac{1}{t-1}\sum_{i=1}^{t-1} \frac{a_i^i - a_i^t}{a_i^i}
    \end{equation}
    where $a_i^j, \forall i \leq j$ represents the BLEU on the $i$-th domain after learning of $j$-th domain\footnote{This definition calculates the average proportion of performance degradation over all previously learned domains, excluding the latest one as it experiences no forgetting.}. This metric is employed to quantify the stability (the ability to prevent forgetting).

    \item SR is defined as:
    \begin{equation}\small
    \operatorname{SR}^t =  1 - \frac{a_t^t}{a_t^M}
    \end{equation}
    where, $a_i^M$ represents the BLEU of $i$-th domain in a mixed-domain training fashion, commonly regarded as the \textbf{upper bound} of CL methods. The saturation rate highlights the system's plasticity (learning ability) when encountering a new task, with a higher rate indicating lower plasticity.
\end{itemize}

\begin{table*}[ht]
\centering
\small
\resizebox{0.95\textwidth}{!}{%
\begin{tabular}{@{}c|lccccccccc@{}}
\toprule
\multirow{2}{*}{Category} & Domain          & General              & IT             & Koran          & Law        & Medical            & Subtitles      & Average        & \multirow{2}{*}{FR{[}\%{]}} &\multirow{2}{*}{Additional storage} \\ \cmidrule(lr){2-9}
                          & Method          & \multicolumn{7}{c}{BLEU}                                                                                                   &              &              \\ \midrule
\multirow{3}{*}{Non-CL}   & Single-domain & 38.00                & \textbf{48.80} & 22.90       & 57.15    & 55.93                   & 32.01          & \textbf{42.47} & -              &  $T \cdot M$              \\
                          & Mixed-domain    & 21.24                & 46.17          & 22.97     & \textbf{60.35}     & \textbf{55.98}  & 29.87          & 39.43          &   -          &  $0$               \\
                          & Adapter         & 38.00                & 44.09          & 22.48   & 53.31 & 51.23                  & \textbf{32.05} & 40.19          &   -         &  $T \cdot A$               \\ \midrule
\multirow{6}{*}{CL}       & Seq-finetune    & 15.81                & 29.29          & 12.16         & 23.90   & 26.50                  & 20.76          & 21.40          & 47.80        &  $0$             \\
                          & EWC             & 24.57                & 36.93          & 17.61         & 46.14  & 43.92                   & 25.01          & 32.36          & 11.47        &  $2M$             \\
                          & KD              & 22.80                & 34.49          & 13.93         & 36.33  & 38.00                 & 24.88          & 28.41          & 32.79          &  $M$           \\
                          & Dynamic-KD      & 27.88                & 31.84          & 14.33           & 40.05  & 39.78                  & 23.53          & 29.57          & 15.33       &  $M$              \\
                          & PTE      & 37.00                & 42.82         & \textbf{{\ul23.06}}           & 52.65  & 49.59                  & {\ul31.69}          & 39.47          & -         &  $T \cdot M[bit] $           \\
                          & F-MALLOC(Ours)            & {\ul \textbf{39.54}} & {\ul 44.33}    &  23.02     & {\ul 53.77}  & {\ul 51.62}      &  31.16    & {\ul 40.57}    & \phantom{0}{\ul 0.71}  &  $T \cdot E$              \\ \bottomrule
\end{tabular}%
}
\caption{BLEU and FR for all domains post-training. The results are averages of 5 different task sequences (Non-CL baselines are independent of task order). The best results are
highlighted in bold. The best CL results are highlighted with an underline. `-' indicates the corresponding methods have no forgetting. The specital tokens denote number of seen tasks($T$), adapter size($A$), model parameter size($M$), binary mask size($M[bit]$) and task embedding size($E$). Note that $M \gg M[bit] > A \gg E$.}
\label{Table:result-avg}
\end{table*}

\begin{table*}[ht]
\centering
\small
\resizebox{0.85\textwidth}{!}{%
\begin{tabular}{@{}lcccccccc@{}}
\toprule
Domain       & Genearl        & IT            & Koran          & Law        & Medical           & Subtitles      & Average        & \multirow{2}{*}{FR{[}\%{]}}  \\ \cmidrule(r){1-8}
Method       & \multicolumn{7}{c}{BLEU}                                                                                            &                             \\ \midrule
Seq-finetune & 21.02          & 23.15         & 11.80          & 31.33          & 36.83          & 30.65          & 25.80          & 37.40                       \\
EWC          & 22.86          & 45.81         & 18.96          & 43.37          & 39.18          & 26.16          & 32.72          & 10.10                       \\
KD           & 25.91          & 29.52         & 13.40          & 40.31          & 44.68          & 31.61          & 30.91          & 27.14                       \\
Dynamic-KD   & 30.35          & 33.83         & 15.56          & 41.65          & 40.90           & 24.62          & 31.15          & 12.22                       \\
PTE          & 37.00          & 43.28         & \textbf{22.98}          & 52.94          & 49.42           & 31.87          & 39.58          & -                  \\
F-MALLOC(Ours)         & \textbf{39.54} & \textbf{44.19} & 22.81 & \textbf{53.64} & \textbf{51.21} & \textbf{31.93} & \textbf{40.55} & \textbf{\phantom{0}0.24}               \\ \bottomrule
\end{tabular}%
}
\caption{BLEU and FR of CL methods for all domains post-training using task sequence 0 (the domain training order corresponds to the sequence in the first row). The best results are highlighted in bold.}
\label{Table:result-seq0}
\end{table*}

\section{Results and Analysis}
Table \ref{Table:result-avg} presents the post-training performances of all nine systems across six domains. Notably, F-MALLOC consistently outperforms all CL baselines on average, with an impressively low forgetting rate of 0.71\%.  In comparison with regularization-based baselines, F-MALLOC demonstrates a better ability to alleviate forgetting and acquire new knowledge. When compared with the SOTA architecture-based method, PTE, F-MALLOC attains higher performance with minimal storage overhead and no prior information about task numbers. 

Regrading the Non-CL baselines, although still trailing behind the upper bound performance, F-MALLOC demonstrates comparable performance to the strong baseline method, Adapter.
These results collectively underscore the effectiveness of F-MALLOC in Continual Learning scenarios for Transformer-based Neural Machine Translation (NMT) systems.

For a more comprehensive analysis and comparison of various CL methods, the following subsections will use task sequence 0: IT $\to$ Koran $\to$ Law $\to$ Medical $\to$ Subtitles as a reference.

\subsection{Comparison with CL methods}

Table \ref{Table:result-seq0} presents the results for task sequence 0. Notably, among the prior CL methods, PTE stands out with the best performance, achieving a BLEU score of 39.58. In contrast, regularization-based methods exhibit inferior performance. The suboptimal results of KD-based approaches (KD and Dynamic-KD) can be attributed to the absence of sample replay in our experimental setting. Without a sample cache from previous tasks, KD struggles to effectively transfer knowledge from the preceding model to subsequent ones. Importantly, F-MALLOC surpasses all CL baselines, delivering the best results in both the BLEU score and forgetting rate.

\paragraph{Robustness against domain order.} A horizontal comparison between Table \ref{Table:result-avg} and Table \ref{Table:result-seq0} for the same method's performance reveals that regularization-base methods such as EWC and KD are sensitive to domain order, resulting in imbalanced performance on the initial and final tasks. In contrast, F-MALLOC exhibits notable resilience to variations in domain order, as evidenced by the balanced performance across different domain orders. This robustness is further substantiated by the low standard deviations presented in Appendix \ref{Appendix:std}.

\begin{figure}[ht]
\centering
\includegraphics[width=0.45\textwidth]{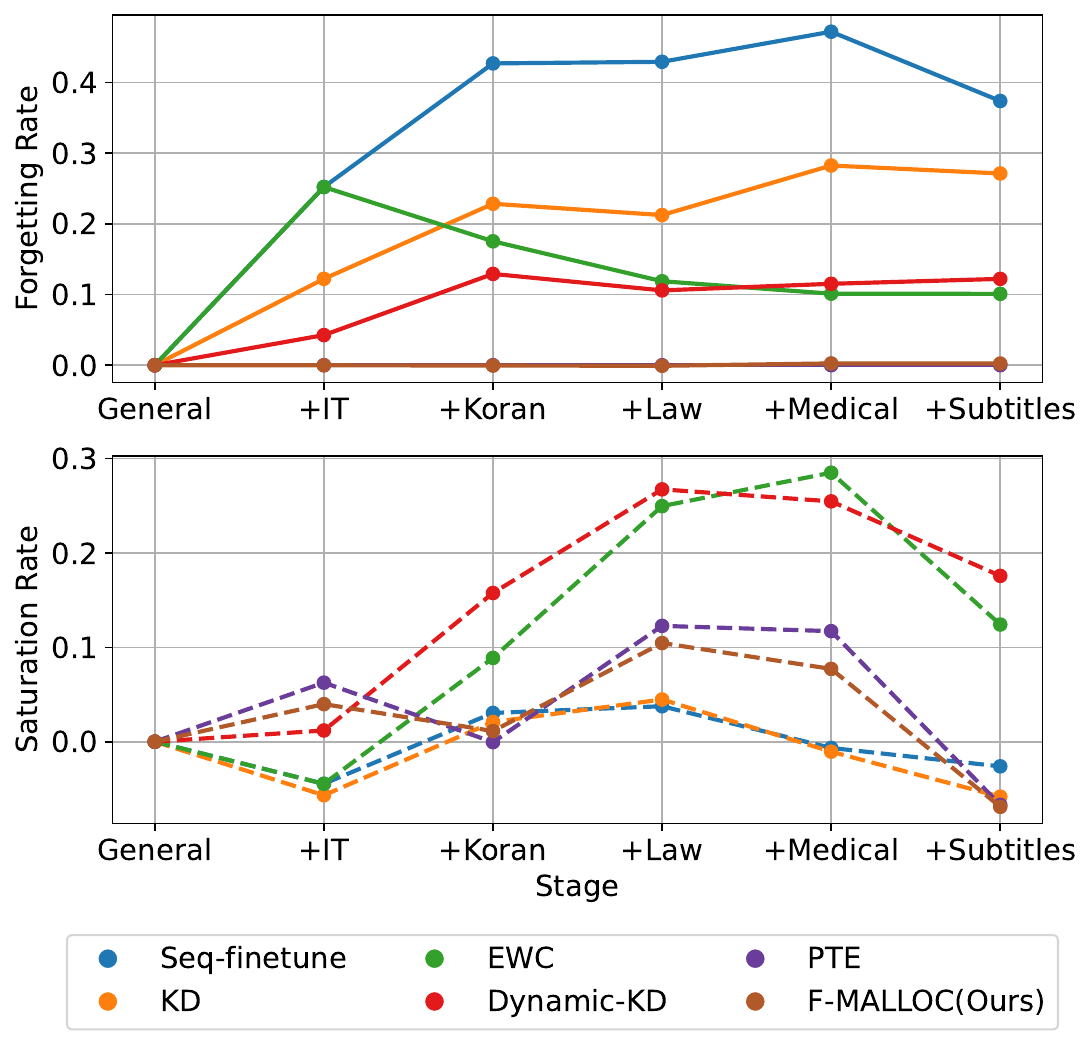}
\caption{Forgetting rate and saturation rate across different training stages.}
\label{Fig:forget_and_saturation}
\end{figure}

\paragraph{Trade-off between stability and plasticity.} 
As depicted in Fig.\ref{Fig:forget_and_saturation}, both EWC and Dynamic-KD exhibit robust abilities to mitigate forgetting. However, they also demonstrate a high saturation rate, suggesting a compromise in their potential to adapt to additional tasks. In contrast, KD achieves a low saturation rate akin to Seq-finetune, but its forgetting rate is notably higher. This observation sheds light on the struggle of regularization-based methods to balance stability and plasticity. Crucially, F-MALLOC excels in both objectives, achieving a harmonious equilibrium between mitigating forgetting and maintaining adaptability.

\subsection{Hyperparameter}

\begin{table}[ht]
\centering
\resizebox{\columnwidth}{!}{%
\begin{tabular}{@{}lcc|lcc@{}}
\toprule
Temp         & BLEU           & FR{[}\%{]}            & Sparsity     & BLEU           & FR{[}\%{]}            \\ \midrule
$\tau_{max}=50$           & 36.83          & 10.81         & $s=0.05$         & 39.09          & 0.40           \\
$\tau_{max}=100$           & 38.33          & \phantom{0}6.89          & $s=0.1$          & 39.87          & 0.15          \\
$\tau_{max}=200$           & 40.22          & \phantom{0}2.37          & $\mathbf{s=0.2}$ & \textbf{40.55} & \textbf{0.24} \\
$\mathbf{\tau_{max}=400}$ & \textbf{40.55} & \textbf{\phantom{0}0.24} & $s=0.3$          & 40.01          & 0.85          \\
$\tau_{max}=800$           & 40.60           & \phantom{0}0.10           & $s=0.4$          & 39.88          & 2.02          \\ \bottomrule
\end{tabular}%
}
\caption{The effect of max temperature $\tau_{max}$ (left) and sparsity $s$ (right). The value used in our experiments is highlighted in bold.}
\label{Table:hyperparameter}
\end{table}

We explored the impact of annealing temperature $\tau_{max}$ and prune sparsity $s$. As outlined in Table \ref{Table:hyperparameter}, a small temperature results in a `soft' mask value, contributing to increased FR. Good results were observed when  $\tau_{max} \geq 400$. Continually increasing the temperature renders the annealing strategy ineffective, resulting in a slower convergence speed. 

In terms of prune sparsity, low sparsity restricts the available capacity for subsequent tasks, while high sparsity adversely affects general domain performance, both contributing to diminished overall performance. Notably, the performance gap across varying prune sparsity levels is relatively small, highlighting the robustness of F-MALLOC.

\subsection{Analyzing Memory Capacity Allocation}



\begin{figure}[ht]
\centering
\includegraphics[width=0.45\textwidth]{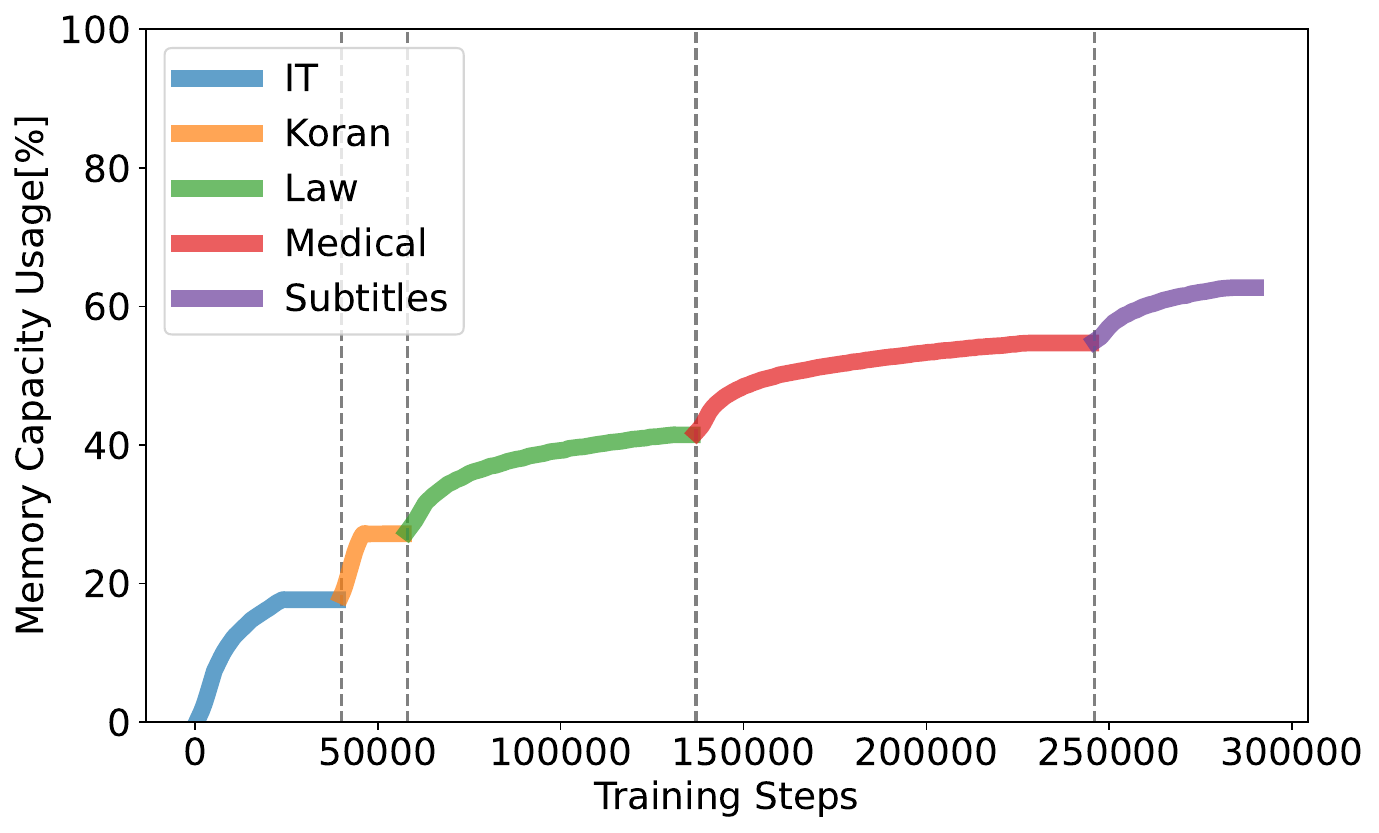}
\caption{Feed-forward memory capacity usage in the training process of task sequence 0. Vertical dash lines indicate task switches.}
\label{Fig:capacity-time}
\end{figure}
F-MALLOC employs a task mask mechanism for the dynamic allocation of feed-forward memories to different tasks. Therefore, by computing the accumulated task mask $m^{<t+1}$ and subsequently binarizing it, we can assess the proportion of allocated memories. As depicted in Fig.\ref{Fig:capacity-time}, the capacity usage undergoes rapid growth in the initial training stage for all tasks, gradually converging at a stable rate thereafter. 

\begin{table}[!ht]
    \centering
    \small
    \begin{tabular}{lccc}
    \toprule
        ~ & \#Token & Init\_loss & Capacity usage[\%] \\ \midrule
        Koran & 0.52M & 3.455 & 12.25 \\ 
        IT & 3.61M & 2.398 & 17.68 \\ 
        Subtitles & 6.25M & 3.144 & 22.48 \\ 
        Medical & 6.90M & 2.284 & 23.15 \\ 
        Law & 19.06M & 1.616 & 22.69 \\ 
    \bottomrule
    \end{tabular}
    \caption{Stastics of data volume (\#Token), task complexity(Init\_loss) and average capacity usage.}
    \label{Table:capacity}
\end{table}

Upon comparing different tasks, we present the statistics of data volume and task complexity for each domain alongside the corresponding average capacity usage in Table.\ref{Table:capacity}. Observing the table, we discern a trend of increased capacity usage with higher data volume. `Law' stands out as an exception to this trend, having three times more tokens yet occupying a similar capacity as `Subtitles' and `Medical'. However, `Law' is evidently the easiest domain, indicated by its lowest initial loss at the start of its training phase. Hence, its low capacity usage is justified by its relatively lower difficulty. Taken together, these observations suggest that our proposed method has acquired a rational and efficient memory allocation strategy, effectively leveraging the data volumes and inherent complexities of the tasks.

Towards the conclusion of the entire training process, approximately 40\% of the feed-forward memory is still `writable'. However, the best-performing baseline, PTE, has already exhausted model capacity. This emphasizes the potential of our proposed method to effectively accommodate additional tasks. 



\subsection{Knowledge Transfer and Domain Similarity from Memory Reusing}
\begin{figure}[ht]
\centering
\includegraphics[width=0.45\textwidth]{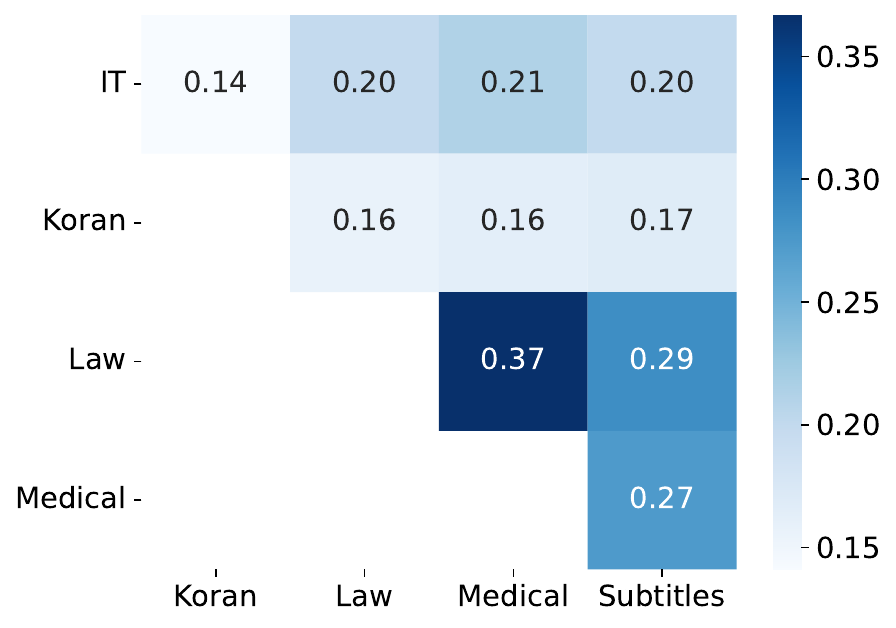}
\caption{Percentage of memory reuse across tasks.}
\label{Fig:capacity-reuse}
\end{figure}

In our proposed method, we employ non-exclusive task masks, allowing feed-forward memories allocated to previous tasks to be reused by subsequent tasks. To investigate the inter-task relationship regarding the allocation of memories, we visually represent the overlap rate among task masks for different tasks. Specifically, we utilize the Jaccard similarity coefficient, defined as $\frac{|m^i \cap m^j|}{|m^i \cup m^j|}$, to assess the memory reuse between task $t_i$ and $t_j$, $i<j$. The results, depicted in Fig.\ref{Fig:capacity-reuse}, reveal a substantial proportion of memory reuse between different tasks. This observation underscores the effectiveness of our non-exclusive masking strategy in facilitating knowledge transfer between tasks.

We further conducted a comparative analysis with the unsupervised domain clustering approach proposed by \citet{DBLP:conf/acl/AharoniG20}. The observed memory reuse rate aligns consistently with domain similarity. Specifically, \citet{DBLP:conf/acl/AharoniG20} found that the `IT' domain cluster attracted the largest number of outliers, primarily from the `Law', `Medical', and `Subtitles' domains. We observed a corresponding pattern of higher memory reuse between these domains in Fig.\ref{Fig:capacity-reuse}. Furthermore, their findings indicated that the `Koran' domain cluster is isolated and attracts the smallest number of outliers, aligning with the notably lower memory reuse observed in our experiments for the `Koran' domain. This alignment highlights the effectiveness of our approach in capturing and leveraging task similarities for improved knowledge transfer.

\section{Related Work}

\paragraph{CL for NMT.} Recent work on CL of NMT can be divided into two categories: regularization-based and architecture-based. Regularization-based techniques address forgetting by incorporating penalty terms to constrain the divergence of model parameters from their previous values. Prominent methods, including Elastic Weight Consolidation (EWC) \citep{DBLP:conf/naacl/ThompsonGKDK19, DBLP:conf/acl/SaundersSGB19} and knowledge distillation \citep{DBLP:conf/mtsummit/DakwaleM17, DBLP:conf/aclnmt/KhayrallahTDK18, DBLP:journals/corr/abs-2212-02800, DBLP:conf/naacl/CaoWCW21}, are widely acknowledged for their effectiveness in the fine-tuning process. \citet{DBLP:conf/emnlp/GuH022} introduced a hard Low Forgetting Risk restriction on all parameters. In contrast to these approaches, our method effectively mitigates forgetting by blocking gradients, showcasing a more efficient strategy. 

Architecture-based methods involve dividing the model into disjoint task-specific components. For instance, \citet{DBLP:conf/naacl/GuFX21} prune the general domain model and subsequently finetune free parameters to adapt to the target domain. Another approach, as demonstrated by \citet{DBLP:conf/aaai/LiangZWQ021}, involves freezing Lottery Ticket Subnetworks to prevent forgetting. Additionally, \citet{DBLP:conf/acl/HuangLMYL23} propose utilizing external models' feed-forward layers and embeddings as a plug-in for knowledge transfer. In comparison to these methods, our approach stands out as it requires no pre-specification of task numbers or space allocation. Moreover, it avoids the need to store an external model or a mask matrix.

\paragraph{Unstructured Pruning for Transformers.} For coarse-grained unstructured pruning of Transformer models, attention-head pruning \citep{DBLP:conf/acl/VoitaTMST19,DBLP:conf/nips/MichelLN19}, layer-dropping \citep{DBLP:conf/iclr/FanGJ20} and block pruning \citep{DBLP:conf/emnlp/LagunasCSR21} have been popularly used. Our proposed pruning method shares similarities with the approach presented in \citet{DBLP:conf/acl/XiaZC22,DBLP:conf/nips/KwonKMHKG22}, where a mask or diagonal matrix is introduced to facilitate pruning. However, our approach diverges in both the estimation of importance and the selection of modules earmarked for pruning..

\section{Conclusion}
This paper introduces F-MALLOC, a pioneering method for CL in NMT systems. By decomposing feed-forward layers into memory cells and implementing a strategic memory allocation approach, F-MALLOC proves effective in simultaneously enhancing new knowledge acquisition and alleviating forgetting. Evaluation with a specialized protocol for CL in NMT, positions F-MALLOC as a superior performer, showcasing substantial improvements, robustness, and extensibility compared to existing approaches. The method's ability to leverage task difficulty and inter-task similarities for enhanced performance represents a significant advancement not seen in previous methods. F-MALLOC not only contribute to the field of CL in NMT but also pave the way for more efficient and adaptable neural network architectures.

\section*{Limitations}
Although our proposed F-MALLOC can effectively alleviate forgetting and exhibits high robustness and extensibility, there are several limitations in our current study: On the one hand, F-MALLOC utilizes a fixed-capacity Transformer, which may limit its capability to adapt to an unrestricted number of tasks. On the other hand, F-MALLOC is designed for domain incremental training. Thus, adding a new language can not be directly solved. We leave these problems for future research.

\section*{Acknowledgements}
We express our gratitude to all the anonymous reviewers for their insightful and valuable comments. We also extend our appreciation to Wenpeng Hu for the fruitful discussions. This work has been supported by the National Natural Science Foundation of China (NSFC) under Grant No. 62206295. 
\bibliography{anthology,custom}

\newpage
\appendix

\section{Dataset Details}
\label{Appendix:dataset}
\begin{table}[ht]
\centering
\small
\begin{tabular}{@{}cc|ccc@{}}
\toprule
\multicolumn{2}{c|}{Dataset}             & Train & Dev.                  & Test                  \\ \midrule
\multicolumn{2}{c|}{WMT17}               & 3.9M  & -                     & -                     \\ \midrule
\multirow{3}{*}{General} & Newstest2019 & -     & -                     & 2000                  \\
                         & Newstest2020 & -     & -                     & 1000                  \\
                         & Newstest2021 & -     & -                     & 785                   \\ \midrule
\multicolumn{2}{c|}{IT}                  & 223K  & \multirow{5}{*}{2000} & \multirow{5}{*}{2000} \\
\multicolumn{2}{c|}{Koran}               & 17K   &                       &                       \\
\multicolumn{2}{c|}{Law}                 & 467K  &                       &                       \\
\multicolumn{2}{c|}{Medical}             & 248K  &                       &                       \\
\multicolumn{2}{c|}{Subtitles}           & 500K  &                       &                       \\ \bottomrule
\end{tabular}%
\caption{Dataset statistics.}
\label{Table:datasets}
\end{table}
Here, we present detailed statistics for the datasets used in our experiments in Table \ref{Table:datasets}, focusing on the translation direction EN $\to$ DE. We employ Moses scripts\footnote{http://www.statmt.org/moses/} for sentence tokenization and truecasing. Additionally, we utilize FastBPE\footnote{https://github.com/glample/fastBPE} to apply Byte Pair Encoding (BPE)\citep{DBLP:conf/acl/SennrichHB16a} to the tokenized data. The dictionary and BPE codes are sourced from the Fairseq WMT19 German-English news translation pretrained model\citep{DBLP:conf/wmt/NgYBOAE19}.

\section{Baseline Details}

\begin{table*}[ht]
\centering
\small
\begin{tabular}{@{}lcccccc@{}}
\toprule
\multirow{2}{*}{Method} & \multicolumn{6}{c}{Domain}                            \\ \cmidrule(l){2-7} 
                        & General & IT    & Koran & Law   & Medical & Subtitles \\ \midrule
Seq-finetune            & 6.28    & 17.62 & 6.35  & 12.91 & 11.69   & 9.20      \\
EWC                     & 5.23    & 5.64  & 1.55  & 4.70  & 6.12    & 5.46      \\
KD                      & 3.97    & 12.85 & 4.91  & 8.69  & 5.74    & 6.16      \\
Dynamic-KD              & 2.59    & 1.81  & 0.87  & 1.85  & 1.49    & 0.76      \\
PTE                     & 0.00    & 0.60  & 0.23  & 0.33  & 0.24    & 0.40      \\
F-MALLOC(Ours)          & 0.00    & 0.71  & 0.27  & 0.62  & 0.61    & 0.57      \\ \bottomrule
\end{tabular}%
\caption{Standard deviation of BLEU score of the proposed F-MALLOC and CL baselines over 5 random task sequences.}
\label{Table:result-std}
\end{table*}

\begin{table*}[ht]
\centering
\small
\begin{tabular}{@{}c|cccccc@{}}
\toprule
\multicolumn{1}{c|}{\multirow{2}{*}{Domain   order}} & \multicolumn{2}{c}{EWC} & \multicolumn{2}{c}{PTE} & \multicolumn{2}{c}{F-MALLOC} \\ \cmidrule(l){2-3}  \cmidrule(l){4-5} \cmidrule(l){6-7}
\multicolumn{1}{c|}{}                                & BLEU     & FR{[}\%{]}   & BLEU     & FR{[}\%{]}   & BLEU       & FR{[}\%{]}      \\ \midrule
IT$\to$Koran$\to$Law$\to$Medical$\to$Subtitles                 & 32.72    & 10.10         & 39.58    & -            & 40.55      & 0.24            \\
Koran$\to$Medical$\to$IT$\to$Law$\to$Subtitles                       & 32.78    & 12.11        & 39.36    & -            & 40.70       & 0.32            \\
Law$\to$IT$\to$Medical$\to$Subtitles$\to$Koran                       & 29.94    & 16.40         & 39.55    & -            & 40.76      & 1.12            \\
Subtitles$\to$Law$\to$Koran$\to$Medical$\to$IT                       & 35.21    & \phantom{0}5.15         & 39.48    & -            & 40.39      & 0.69            \\
Medical$\to$Law$\to$Koran$\to$Subtitles$\to$IT                       & 31.16    & 13.59        & 39.37    & -            & 40.47      & 1.17            \\ \bottomrule
\end{tabular}%
\caption{Result in different domain orders. The best-performing regularization-based baseline, EWC, and architecture-based baseline, PTE, were chosen for comparison.}
\label{FullResult}
\end{table*}

\label{Appendix:baseline}
\paragraph{Non-Continual Learning Methods.} Each of these baselines constructs a distinct model (or module) for each task independently. Consequently, they do not encounter CF and lack knowledge transfer between tasks.
\begin{itemize}
    \item \textbf{Single-domain} continues to train the general domain model on target domain data, respectively. 
    \item \textbf{Mixed-domain} trains the general domain model on combined multi-domain data, which is considered the \textbf{upper bound} of CL methods.
    \item \textbf{Adapter \citep{DBLP:conf/emnlp/BapnaF19}} inserts adapters on each transformer block of the general domain model as proposed by \citet{DBLP:conf/emnlp/BapnaF19}. We set the bottleneck dimension to 64 and only finetune the adapters.
\end{itemize}
\paragraph{Continual Learning Methods:} 
\begin{itemize}
    \item \textbf{Sequential Fine-tuning} continues to train the general domain model on target domains sequentially, without incorporating any mechanism to address CF.
    \item \textbf{Elastic Weight Consolidation (EWC) \citep{DBLP:conf/naacl/ThompsonGKDK19,DBLP:conf/acl/SaundersSGB19}} is a popular regularization-based CL method that adopts elastic weights consolidation to introduce $L_2$ regularization, penalizing parameter changes. The training objective is:
    \begin{equation*}
        \mathcal{L}_{\text{EWC}}(\theta) = \mathcal{L}_{\text{CE}}(\theta) + \alpha \sum_{i} F_{i}(\theta_i - \theta_i^G)^2
    \end{equation*}
    In this equation, $\theta$ represents the model parameters, $F$ is the diagnosis of the Fisher information matrix, and $\alpha$ is a hyperparameter controlling the strength of regularization. To extend this method to a multi-stage scenario, we adopt the accumulated Fisher information, as proposed by \citet{DBLP:journals/corr/abs-1712-03847}.
    \item \textbf{Knowledge Distillation (KD) \citep{DBLP:conf/aclnmt/KhayrallahTDK18,DBLP:conf/mtsummit/DakwaleM17}} introduces a regularization (reg) term into the training objective. The reg term is formulated in the spirit of knowledge distillation, minimizing the cross-entropy between the original (teacher) model's output distribution and that of the new (student) model. A hyperparameter $\alpha$ is introduced to interpolate the reg term and the NLL loss.
    \begin{equation*}
        \mathcal{L}_{\text{EWC}}(\theta) = \mathcal{L}_{\text{CE}}(\theta) + \alpha \mathcal{L}_{\text{KD}}(\theta)
    \end{equation*}
    In our experiments, the weight of the KD term is set to 0.1.
    \item \textbf{Dynamic Knowledge Distillation (Dynamic-KD) \citep{DBLP:conf/naacl/CaoWCW21}} propose dynamically adjusting the weight of KD loss to better alleviate CF in a multi-stage CL scenario. The bias correction module is omitted due to its incompatibility with the pretrained model.
    \item \textbf{Prune Then Expand (PTE) \citep{DBLP:conf/naacl/GuFX21}} employs unstructured pruning to trim the general domain model, followed by training the pruned parameters for the target domain. In the context of multi-stage CL, we uniformly distribute the pruned parameters across all subsequent tasks.
\end{itemize}

\section{Implementation Details}

\label{Appendix:implementation_details}
\paragraph{pretrained Model.} All methods are implemented with the Fairseq toolkit \citep{DBLP:conf/naacl/OttEBFGNGA19}. We adopt the WMT’19 German-English news translation task winner \citep{DBLP:conf/wmt/NgYBOAE19} as the pretrained general domain model. It is a Transformer encoder-decoder model \citep{DBLP:conf/nips/VaswaniSPUJGKP17} with 6 layers, 1,024-dimensional representations, 8,192-dimensional feed-forward layers, and 8 attention heads. Apart from WMT’19 training data, this model is trained on over 10 billion tokens of backtranslation data and finetuned on the Newstest test sets from years before 2018. In our experiments, we do not use ensembles or n-best reranking.

\paragraph{Hyper-parameters.} Unless explicitly stated otherwise, consistent hyperparameters are applied across all experiments. We utilize the Adam optimizer \citep{DBLP:journals/corr/KingmaB14} with the same learning rate scheduler as detailed in \citet{DBLP:conf/nips/VaswaniSPUJGKP17}. The learning rate is set to 1e-4 for all systems during the fine-tuning process. Training is stopped when there is no performance improvement for 5 consecutive validation steps.

For inference, we employ beam search with a beam size of 5 for all systems. The default parameter of BLEU is utilized in evaluation.

All experiments are done on 8 NVIDIA RTX 3090 GPUs.

\section{Standare Deviations}
\label{Appendix:std}

This section reports the standard deviations of the results in Table \ref{Table:result-avg}. We only include the CL baselines here, since Non-CL baselines‘ performance is independent of the domain order. As shown in Table \ref{Table:result-std}, F-MALLOC achieves low standard deviations, indicating its robustness.

\section{Result in Different Domain Orders}

Table \ref{FullResult} shows the performance of F-MALLOC along with two strong baselines, EWC and PTE, in other domain orders. F-MALLOC outperforms both EWC and PTE, highlighting its efficacy across different domain order scenarios.

\end{document}